\documentclass{article}

\usepackage{microtype}
\usepackage{graphicx}
\usepackage{subfigure}
\usepackage{booktabs} % for professional tables
\usepackage{colortbl}
\usepackage{pifont}
\usepackage{hyperref}
\usepackage{multirow}

\usepackage[accepted]{icml2024}

\usepackage{amsmath}
\usepackage{amssymb}
\usepackage{mathtools}
\usepackage{amsthm}

\usepackage[capitalize,noabbrev]{cleveref}

\theoremstyle{plain}

\theoremstyle{definition}

\theoremstyle{remark}

\usepackage[textsize=tiny]{todonotes}

\icmltitlerunning{SAM as the Guide: Mastering Pseudo-Label Refinement in Semi-Supervised
Referring Expression Segmentation}

\begin{document}

\twocolumn[
\icmltitle{SAM as the Guide: Mastering Pseudo-Label Refinement in Semi-Supervised
Referring Expression Segmentation}



\icmlsetsymbol{equal}{*}

\begin{icmlauthorlist}
\icmlauthor{Danni Yang}{equal,xmu}
\icmlauthor{Jiayi Ji}{equal,xmu}
\icmlauthor{Yiwei Ma}{xmu}
\icmlauthor{Tianyu Guo}{xmu}
\icmlauthor{Haowei Wang}{xmu,tencent}
\icmlauthor{Xiaoshuai Sun}{xmu}
\icmlauthor{Rongrong Ji}{xmu}
\end{icmlauthorlist}

\icmlaffiliation{xmu}{Key Laboratory of Multimedia Trusted
Perception and Efficient Computing, Ministry of Education
of China, School of Informatics, Xiamen University, 361005,
P.R. China.}
\icmlaffiliation{tencent}{Youtu Lab, Tencent, Shanghai, China}

\icmlcorrespondingauthor{Xiaoshuai Sun}{xssun@xmu.edu.cn}

\icmlkeywords{Machine Learning, ICML}

\vskip 0.3in
]



\printAffiliationsAndNotice{\icmlEqualContribution} % otherwise use the standard text.

\begin{abstract}
In this paper, we introduce SemiRES, a semi-supervised framework that effectively leverages a combination of labeled and unlabeled data to perform RES. A significant hurdle in applying semi-supervised techniques to RES is the prevalence of noisy pseudo-labels, particularly at the boundaries of objects. SemiRES incorporates the Segment Anything Model (SAM), renowned for its precise boundary demarcation, to improve the accuracy of these pseudo-labels. Within SemiRES, we offer two alternative matching strategies: IoU-based Optimal Matching (IOM) and Composite Parts Integration (CPI). These strategies are designed to extract the most accurate masks from SAM's output, thus guiding the training of the student model with enhanced precision. 
In instances where a precise mask cannot be matched from the available candidates,  we develop the Pixel-Wise Adjustment (PWA) strategy, guiding the student model's training directly by the pseudo-labels. 
Extensive experiments on three RES benchmarks—RefCOCO, RefCOCO+, and G-Ref reveal its superior performance compared to fully supervised methods. 
Remarkably, with only 1\% labeled data, our SemiRES outperforms the supervised baseline by a large margin, e.g. +18.64\% gains on RefCOCO val set. 
The project code is available at \url{https://github.com/nini0919/SemiRES}.
\end{abstract}

\begin{figure}[t] 
\centering 
\includegraphics[width=1.0\columnwidth]{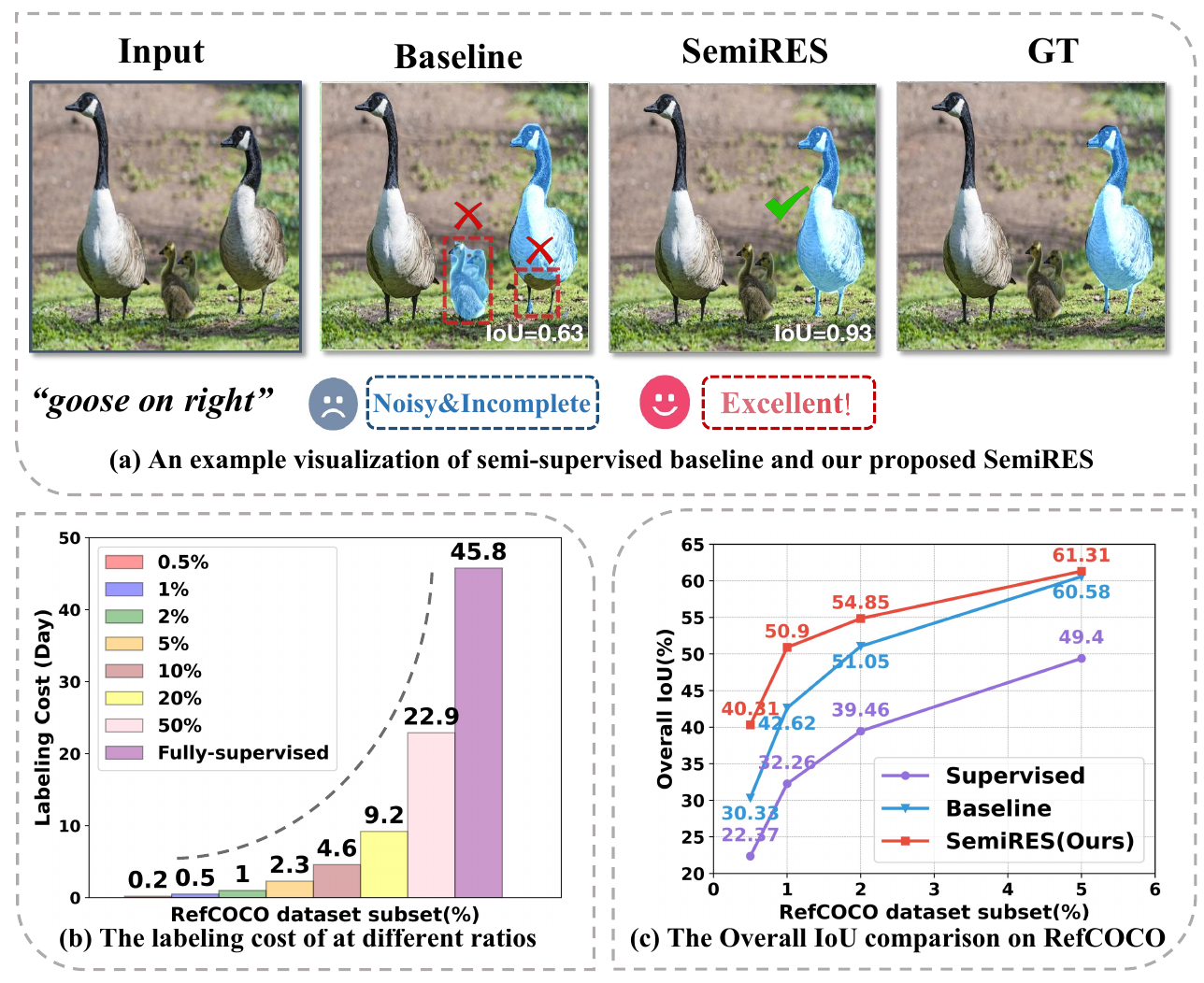}  
%\vspace{-2em}
\caption{%Quantitative and qualitative comparison and annotation cost analysis: (a) An example about absolute position. The result of baseline is noisy and incomplete, whereas our proposed SemiRES effectively captures semantic information, providing accurate segmentation. (b) Histogram of annotation costs for semi-supervised training under different labeled ratios on RefCOCO. (c) Quantitative performance comparison of supervised model, baseline and SemiRES under different labeled ratios on RefCOCO.
(a) A large number of noisy and incomplete cases exist in pseudo-labels. The proposed SemiRES can address this issue.
(b) Analysis shows labeling a small portion of RefCOCO data can greatly reduce costs.
(c) Our method substantially improves performance,  even with a small number of annotated samples.
} 
\label{fig:intro}
%\vspace{-1.5em}
\end{figure}

\section{Introduction}
\label{submission}
Referring Expression Segmentation (RES) has attracted considerable attention from the fields of vision and language research~\cite{hu2016segmentation,chen2019referring,huang2020referring,liu2019learning,liu2023polyformer}. Unlike common visual grounding tasks such as phrase localization~\cite{bajaj2019g3raphground,chen2017query,dogan2019neural,plummer2015flickr30k,plummer2017phrase} and referring expression comprehension~\cite{yang2020improving,yu2018mattnet,deng2021transvg,kamath2021mdetr,huang2021look}, RES requires precise pixel-level segmentation of an object within an image, as directed by a referring expression, which goes beyond simple bounding box identification.

Despite advancements, the high demand for labeled data presents a significant barrier to the deployment of RES, particularly in domains where labeling is prohibitively expensive, such as medical imaging and autonomous driving. The labor intensity of the task is underscored by findings from \cite{kim2023devil}, indicating that an average mask segmentation requires approximately 79.1 seconds. This is further exemplified by the extensive annotation efforts needed for benchmark datasets like RefCOCO~\cite{yu2016modeling}, RefCOCO+~\cite{yu2016modeling}, and G-Ref~\cite{mao2016generation,nagaraja2016modeling}, which consist of tens of thousands of labeled instances and necessitate substantial time investments, as illustrated in Fig.~\ref{fig:intro} (b). The cost and time implications, alongside the potential for inaccuracies in manual labeling, present considerable challenges for the scalability and reliability of RES models, highlighting the urgent need for more efficient methodologies.

To address the challenges outlined above, we pioneer a semi-supervised learning framework tailored for RES, utilizing a combination of a small subset of image-text pairs with segmentation annotations and a large corpus of unannotated pairs to train the model. This approach has been widely validated across the fields of computer vision~\cite{sohn2020fixmatch,olsson2021classmix,chen2023softmatch}, natural language processing~\cite{miyato2016adversarial,cheng2019semi,zou2023jointmatch}, and vision-language research~\cite{yang2023semi,sun2023refteacher,jin2023refclip}. Yet, its application in RES has not yet been explored. We establish a baseline for semi-supervised RES, encompassing a comprehensive pipeline that goes beyond data augmentation~\cite{zhang2017mixup,hendrycks2019augmix,cubuk2019autoaugment,zhao2023augmentation} and Exponential Moving Average~\cite{kingma2014adam,he2020momentum,grill2020bootstrap,tarvainen2017mean} training mechanisms. However, this baseline encounters a substantial challenge: pseudo-labels are significantly noisy, particularly at the edges of instances, which can trap the model in suboptimal performance, as depicted in Fig.~\ref{fig:intro} (a). The crux of semi-supervised learning lies in refining these pseudo-labels to enhance their quality. Previous methods have addressed this by employing confidence-based pseudo-label filtering strategies~\cite{sohn2020fixmatch} or auxiliary correcting networks~\cite{kwon2022semi,kim2023devil}. While intuitively appealing, relying solely on confidence for filtering may lead to the under-utilization of unlabeled data and lack flexibility in handling diverse noise in pseudo-labels.

To tackle the aforementioned issues, we introduce a novel semi-supervised RES framework named SemiRES. The motivation behind SemiRES is to leverage the robust segmentation capabilities of SAM~\cite{kirillov2023segment} to rectify pseudo-labels, especially around the edges of instances. Specifically, we employ SAM to extract multi-scale masks from original images to build a proposal library. The central concept of SemiRES is to retrieve one or multiple proposals from this library to reconstruct pseudo-labels. To achieve this, we propose two alternative strategies: IoU-based Optimal Matching (IOM) and Composite Parts Integration (CPI). The first assumes that the proposal library contains masks closely approximating the target instance, thus utilizing IoU to directly identify and replace the pseudo-label with the most corresponding mask from the library. The second strategy moves away from this assumption, instead using the pseudo-label to select different part-specific proposals from the library to assemble a complete mask. In cases where a suitable replacement cannot be retrieved from the proposal library, we default to optimizing the student model using the pseudo-label itself. To enhance training in such scenarios, we have devised a Pixel-Wise Adjustment (PWA) strategy that adjusts the final loss on a per-pixel basis according to the confidence levels on the pseudo-label. 

To further qualitatively validate the effectiveness of our proposed SemiRES, we conduct extensive experiments on three RES benchmark datasets—RefCOCO, RefCOCO+, and G-Ref. 
Our experiments show that SemiRES notably surpasses both supervised and semi-supervised baselines in all settings, for example, gaining +18.64\% and +8.28\% on 1\% labeled RefCOCO as shown in Fig.~\ref{fig:intro} (c), which highlights its significant real-world application potential.

To sum up, the contributions of this paper are three-fold:
\begin{itemize}
    \item We first present SemiRES, a semi-supervised framework tailored for RES that efficiently trains models using a minimal amount of labeled data, thus reducing dependence on expensive pixel-level annotations.
    \item We introduce two alternative strategies, IoU-based Optimal Matching (IOM) and Composite Parts Integration (CPI), that leverage the SAM's edge-segmentation proficiency to produce superior-quality pseudo-labels.
    \item Our SemiRES framework achieves notable performance improvements on three benchmark datasets RefCOCO, RefCOCO+, and G-Ref, demonstrating significant gains in model accuracy while concurrently cutting down on labeling costs.
\end{itemize}

\section{Related Work}
\subsection{Referring Expression Segmentation}
        Referring Expression Segmentation ~\cite{chen2019referring,huang2020referring,liu2023polyformer,liu2023gres,hu2023beyond,kim2023shatter,yang2022lavt} is a multimodal task involving both image segmentation and natural language understanding. It aims to identify specific target regions within an image according to natural language expressions. In recent years, RES methods have made significant progress, with Transformer-based backbones emerging as the predominant choice for this task. Additionally, RES methods can be categorized into two main types: one-stage and two-stage approaches. One-stage methods~\cite{chen2019see,hu2020bi,hui2020linguistic} usually use end-to-end networks for prediction, while two-stage methods~\cite{yu2018mattnet,liu2022instance} employ an instance segmentation network to generate a set of instance proposals before selecting the target instance. Recently, GRES~\cite{liu2023gres}introduces multi-target and no-target expressions, extending the classic RES to refer to an arbitrary number of target objects.\par

\subsection{Semi-Supervised Semantic Segmentation}
Semi-supervised learning~\cite{french2019semi,xu2021dash, wu2021semi,olsson2021classmix,jiang2022maxmatch,yang2023semi}, using a small amount of labeled data alongside a larger pool of unlabeled data for training, has widespread applications in computer vision and natural language processing. 
%For instance, manually annotating pixel-level annotations for semantic segmentation is a laborious and expensive process. Therefore, leveraging easily accessible unlabeled data holds significant value. 
In recent years, semi-supervised semantic segmentation has developed rapidly, with many new research works emerging~\cite{ouali2020semi,chen2021semi,wang2022semi,yang2022revisiting}. Building upon FixMatch~\cite{sohn2020fixmatch}, PseudoSeg~\cite{zou2020pseudoseg} extends weak consistency to strong consistency in segmentation scenarios and incorporates a calibration module for pseudo-label refinement. ReCo~\cite{liu2021bootstrapping} samples classes prone to confusion across all categories to assist the segmentation network for better representations. AugSeg~\cite{zhen23augseg}, a simple yet effective method, primarily enhances the performance of semi-supervised semantic segmentation through data augmentation.\par

\subsection{Segment Anything Model}
The recent Segment Anything Model (SAM)~\cite{kirillov2023segment} has made significant advancements in pushing the boundaries of segmentation, greatly boosting the development of foundational models for computer vision. %SAM is a powerful segmentation model comprised of a large image encoder, a prompt encoder, and a lightweight mask decoder.
Trained on SA-1B of over 1 billion masks, it aims to segment any object in any given image without requiring any additional task-specific adaptation. SAM is proved to be capable of solving various tasks, such as medical image analysis~\cite{ma2023segment,shi2023generalist},  adversarial attacks~\cite{guan2023badsam,zhang2023attack},  image inpainting~\cite{yu2023inpaint}, image editing~\cite{xie2023everything}, image captioning~\cite{wang2023caption}. Recently, several works~\cite{zhang2023personalize,liu2023matcher} have incorporated SAM for one-shot learning, contributing significantly to SAM's multifaceted development. In this paper, we investigate how to leverage SAM to enhance pseudo-labels and improve the performance of semi-supervised learning.

\section{Method}
\label{sec:method}

\subsection{Task Definition}
Before diving into SemiRES, we begin with illustrating the task definition of semi-supervised referring expression segmentation (RES). We usually have a small labeled dataset $\mathcal{D}_l=\left\{\left(\left(\mathcal{I}_i^l,\mathcal{T}_i^l\right),Y_i^l\right)\right\}_{i=1}^{N^l}$ and a much larger unlabeled dataset $\mathcal{D}_u=\left\{\left(\left(\mathcal{I}_i^u, \mathcal{T}_i^u\right),\varnothing \right)\right\}_{i=1}^{N^u}$, where $\mathcal{I}_i^l, \mathcal{I}_i^u$ denote the $i$-th labeled and unlabeled image, respectively; $\mathcal{T}_i^l$ and $ \mathcal{T}_i^u$ are the corresponding language expressions; ${N^l}$ and ${N^u}$ are the number of labeled and unlabeled data, with ${N^l} \ll {N^u}$. 
It is crucial to note that the unlabeled set $\mathcal{D}_u$ lacks ground truth mask labels, utilizing language expressions solely as input. 
Our primary aim is to leverage this small labeled set alongside a large unlabeled set to achieve competitive performance in the RES task.
\subsection{Semi-Supervised Baseline}
\label{baseline}
We introduce a semi-supervised baseline for RES based on a teacher-student network structure. 
This approach unfolds in two stages:\par
\noindent\textbf{Stage1: Burn-In Stage.}
In the semi-supervised teacher-student framework, achieving proper parameter initialization is crucial for accelerating the convergence of training during the mutual learning stage~\cite{liu2021unbiased}. During the Burn-In stage, we train the pre-trained model using only labeled data. The optimization objective is defined as follows:
\begin{equation}
\mathcal{L}_{sup}=\frac{1}{H \times W} \sum_{j=1}^{H \times W} \mathcal{L}_{BCE}\left(M_{i,j}^l,Y_{i,j}^l\right),
\label{1}
% \vspace{-0.5em}
\end{equation}
where $M_{i,j}^l$ denotes the prediction mask of Burn-In model for the $j$-th pixel of $i$-th labeled image, $Y_{i,j}^l$ denotes the corresponding ground truth, $\mathcal{L}_{BCE}$ denotes binary cross entropy loss~\cite{csiszar2008axiomatic}.
\textbf{\begin{figure*}[]
  \centering
\includegraphics[width=1\linewidth]{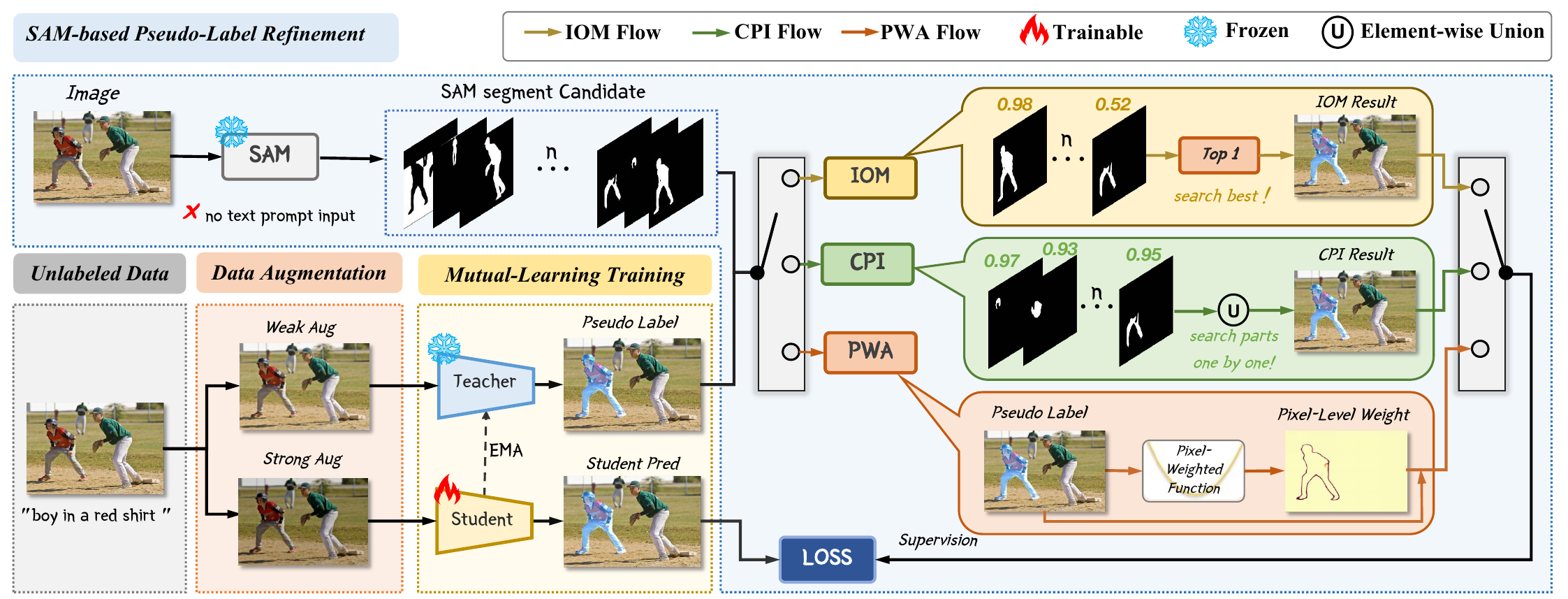}
  % \vspace{-0.8em}
  \caption{%An overview of our proposed SemiRES.
  An overview of the proposed SemiRES, featuring a teacher-student network with data augmentation and mutual learning. It includes SAM-based pseudo-label refinement using IOM or CPI strategies, and PWA supervision when matches are not found.
  }
  \label{fig:framework}
  % \vspace{-1.5em}
\end{figure*}}
\\
\noindent\textbf{Stage2: Mutual-Learning Stage.}
After the Burn-In stage, we use the trained weights $\theta$ to initialize both the teacher and student models. This process is defined as follows:
% \vspace{-0.7em}
\begin{equation}
\theta_t \leftarrow \theta,  \theta_s \leftarrow \theta,
\label{2}
% \vspace{-0.5em}
\end{equation}
where $\theta_t, \theta_s, \theta$ denote the parameters of the teacher, student and Burn-In model, respectively.\par
During the mutual learning stage, the teacher generates pseudo-labels for unlabeled data to supervise the training of the student, which is defined as follows:
% \vspace{-0.7em}
\begin{equation}
\mathcal{L}_{unsup}=\frac{1}{H \times W} \sum_{j=1}^{H \times W} \mathcal{L}_{BCE}\left( M_{i,j}^u,\hat M_{i,j}^u\right),
\label{1}
% \vspace{-0.5em}
\end{equation}
where $M_{i,j}^u$ and $\hat M_{i,j}^u$ denote the predicted mask for $j$-th pixel of $i$-th unlabeled image by student and teacher, respectively. 

Simultaneously, the student continues to train on a small subset of labeled data, jointly optimizing with these two components of loss function, which is defined as follows:
% \vspace{-0.7em}
\begin{equation}
\mathcal{L}=\lambda_{sup}\mathcal{L}_{sup}+\lambda_{unsup} \mathcal{L}_{unsup},
\label{eq:unsupervised}
% \vspace{-0.3em}
\end{equation}
where $\lambda_{sup}$ and $\lambda_{unsup}$ is the hyperparameter of supervised loss $\mathcal{L}_{sup}$ and unsupervised loss $\mathcal{L}_{unsup}$.

To maintain the stability of pseudo-labels, we forego gradient backpropagation for updating the teacher model's parameters. Instead, we employ the Exponential Moving Average (EMA) method to create an aggregated model reflecting both the current and previous states. EMA's effectiveness has been substantiated in numerous studies~\cite{kingma2014adam,ioffe2015batch,he2020momentum,grill2020bootstrap,tarvainen2017mean}. The use of EMA not only improves the teacher model's accuracy but also its stability, making it a valuable tool during the mutual learning stage, which is formulated as follows:
\begin{equation}
\theta_t \leftarrow \alpha \theta_{t}+(1-\alpha) \theta_s,
\label{ema}
% \vspace{-0.3em}
\end{equation}
where $\alpha$ is the decay coefficient of EMA, typically set within the small range of 0.9 to 0.999.\par
\begin{figure}[h] 
\centering 
\includegraphics[width=0.85\columnwidth]{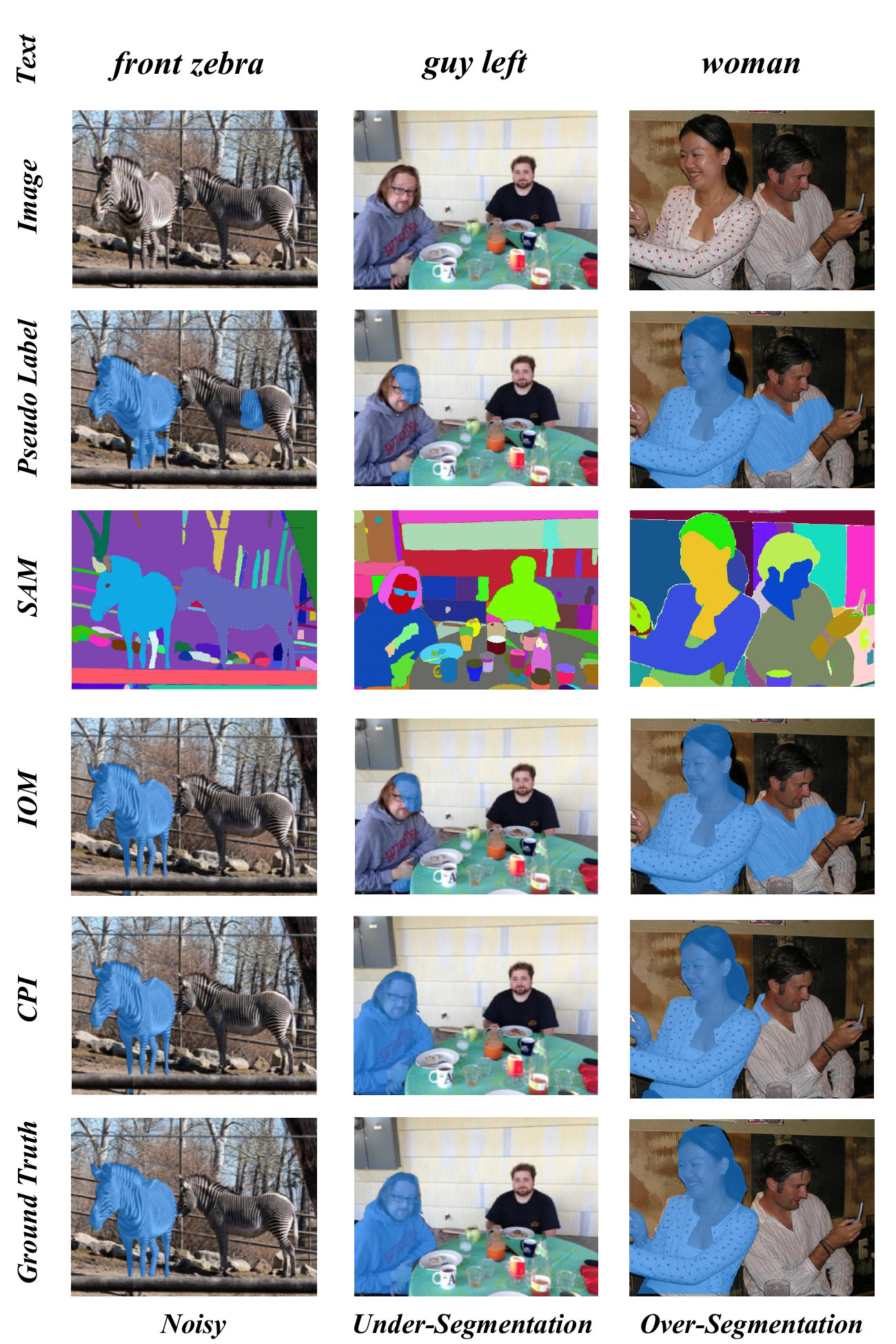} 
% \vspace{-1em}
% \caption{Demonstration of how IOM and CPI work.} 
\caption{Visualization of the principles behind IOM and CPI addressing pseudo-label issues in different cases.}
\label{fig3}
% \vspace{-2em}
\end{figure}
\subsection{The Proposed SemiRES}
\subsubsection{Overview}
The overview of our proposed SemiRES is depicted in Fig.~\ref{fig:framework}. SemiRES inherits the semi-supervised framework introduced in  Sec.~\ref{baseline} and proposes new strategies to address the challenges of noisy pseudo-labels encountered by regular semi-supervised frameworks, which limit the extraction of knowledge from unlabeled data. The core idea is to exploit the powerful edge segmentation capabilities of SAM. The central question of our research is how to utilize these masks to refine the noisy pseudo-labels. In this paper, we propose two alternative matching strategies, IoU-based Optimal Matching (IOM) and Composite Parts Integration (CPI), to select masks that contribute to the final pseudo-labels, as detailed in  Sec.~\ref{SAM_Match}. Moreover, when segments generated by SAM cannot be matched with pseudo-labels, we introduce a Pixel-Wise Weighted Adjustment (PWA) scheme to focus the model on more reliable pixels, thereby improving performance, as outlined in  Sec.~\ref{PWLA}.

\subsubsection{SAM-based Pseudo-Label Refinement}
\label{SAM_Match} 
Despite SAM's powerful segmentation capabilities, effectively harnessing these for pseudo-label refinement is an area ripe for investigation. We have formulated two strategies for matching SAM-generated segments with the original pseudo-labels to improve their accuracy. Before deploying these strategies, we utilize SAM's ``Segment Everything'' feature to create an extensive proposal library of multi-scale candidate segments for our dataset offline, eliminating the need for specific prompts. To optimize storage space, we implement the Run Length Encoding (RLE) algorithm\footnote{\url{https://en.wikipedia.org/wiki/Run-length_encoding}}. Considering that SAM is capable of producing hundreds to thousands of intricate segments per image, adopting efficient storage solutions is crucial. Importantly, while the RLE algorithm achieves high compression rates, it also preserves the precision of the candidate masks.\par
\textbf{IoU-based Optimal Matching (IOM).} 
To achieve our goal, we initially consider a more straightforward approach, premised on the robust multi-scale segmentation ability of the Segment Anything Model (SAM). We hypothesize that the proposal library, constructed as previously mentioned, likely contains a close approximation of the ideal target segmentation. Therefore, our task simplifies devising a method to retrieve this optimal mask from the library.
Our method involves an IoU-based selection process, which computes the similarity between the pseudo-labels and each segment generated by SAM. We aim to identify the segment with the highest similarity score, ensuring it aligns closely with the overall target mask. 
The similarity measure is calculated using the Intersection over Union (IoU) metric, a standard in object detection and segmentation tasks that quantifies the extent of overlap between two areas. By selecting the segment with the top-1 IoU score from the candidate pool, we can effectively align our model's output with the most accurate representation of the intended segmentation, as detailed below:
% \vspace{-0.7em}
\begin{equation}
s^k = \frac{\sum_{j=1}^{H \times W}\left(\hat {M}_{i,j}^u \cap \widetilde{M}_{i,j,k}^u\right)}{\sum_{j=1}^{H \times W}\left(\hat {M}_{i,j}^u \cup \widetilde{M}_{i,j,k}^u \right)},
\label{6}
% \vspace{-0.5em}
\end{equation} 
\noindent where $\hat M_{i,j}^u$ and $\widetilde M_{i,j,k}^u$ denote the pseudo-labels and $k$-th segment mask generated by SAM for $j$-th pixel of $i$-th unlabeled image $\mathcal{I}_i^u$, respectively. When the score $s^k$ exceeds a certain threshold $IoU_{rate}$, the matched mask will replace the pseudo-label. The detailed  matching schemes can be found in Algorithm~\ref{al:alg}, particularly within lines 4 to 9.\par
\textbf{Composite Parts Integration (CPI).}
In our exploration of semi-supervised referring expression segmentation, we recognize that while the IoU-based Optimal Matching (IOM) strategy is generally straightforward and effective, it may falter in certain scenarios. One such instance occurs when the proposal library lacks an ideal target segmentation, rendering even the most sophisticated matching algorithm incapable of finding an appropriate guide mask. Another instance is when the disparity between the pseudo-labels and the desired segmentation is too substantial to allow for effective correction. We have noted that the original pseudo-labels generated by the teacher model can suffer from either under-segmentation or over-segmentation of the target instances, as depicted in Fig.~\ref{fig3}. These inaccuracies diminish the quality of the pseudo-labels, providing erroneous guidance to the student model and impeding its learning. 

Under-segmentation is characterized by incomplete coverage of the target instance, missing activation for certain region pixels. To address this, we aim to identify larger regions within the proposal library to rectify the pseudo-labels. Our selection is based on the overlap ratio with the pseudo-labels, calculated as follows:
\begin{equation}
s_1^k = \frac{ \sum_{j=1}^{H \times W} \left(\hat{M}_{i,j}^u \cap \widetilde{M}_{i,j,k}^u\right)}{\sum_{j=1}^{H \times W} \left(\hat{M}_{i,j}^u\right) + \epsilon},
\label{7}
\end{equation} 
where $\epsilon$ is the smoothing factor to prevent a denominator of zero. When the overlap ratio $s_1^k$ exceeds a predefined threshold $inter_1$, the $k$-th segment $\widetilde{M}_{i,:,k}^u$ generated by SAM is selected and subsequently merged to replace the pseudo-labels. This method is referred to as Composite Parts Integration for Under-segmentation (CPI-U). Conversely, over-segmentation introduces erroneous regions into the segmentation. To mitigate this, we seek to leverage SAM's segmentation to filter out the extraneous noise. The selection is based on the overlap ratio with the candidate mask, computed as:
% \vspace{-0.8em}
\begin{equation}
s_2^k = \frac{ \sum_{j=1}^{H \times W} \left(\hat {M}_{i,j}^u \cap \widetilde{M}_{i,j,k}^u\right)}{\sum_{j=1}^{H \times W}\left(\widetilde{M}_{i,j,k}^u\right)}.
\label{8}
% \vspace{-0.5em}
\end{equation}
Likewise, when the ratio $s_2^k$ is above the set threshold $inter_2$, the segment $\widetilde{M}_{i,:,k}^u$ generated by SAM is chosen and integrated to refine the pseudo-labels. This approach is termed Composite Parts Integration for Over-segmentation (CPI-O). When both conditions are met, we form the overarching CPI strategy. The detailed matching schemes can be found in Algorithm~\ref{al:alg}, particularly within lines 10 to 15.\par
\begin{algorithm}[t]
\caption{Pseudo code for our proposed SemiRES}
\label{al:alg}
\textbf{Input:} Teacher's predicted pseudo mask $\hat {M}_{i}^u$ for $i$-th unlabeled image $\mathcal{I}_i^u$, the multi-scale mask $\{\widetilde{M}_{i,:,k}^u\}_{k=1}^{N_i}$ generated by SAM, the number of masks $N_i$ generated by SAM  for $\mathcal{I}_i^u$,  selected strategy $S$, current top-1 score $s_{top}$\\
\textbf{Output:} Enhanced pseudo mask $\ddot M_{i}^u$ \par
\begin{algorithmic}[1]
\STATE Initialize $\ddot{M}_{i}^u \leftarrow \emptyset, s_{top} \leftarrow 0$;
\FOR{k in $1...N_i$}
\STATE Get $k$-th SAM's segment $\widetilde{M}_{i,:,k}^u$ for image $\mathcal{I}_i^u$;
\IF{$S$ == ``IOM''}{
\STATE Compute the score $s^k$ by Eq.(\ref{6});
\IF{$s^k > IoU_{rate}$ and $s^k> s_{top}$}{
\STATE $\ddot M_{i}^u \leftarrow \widetilde{M}_{i,:,k}^u$ ,
$s_{top} \leftarrow s$};
\ENDIF
}
\ENDIF
\IF{$S$ == ``CPI''}{
\STATE Compute the score $s^k_1$, $s^k_2$ by Eq.(\ref{7}) and Eq.(\ref{8});
\IF{$s^k_1 > inter_1$ or $s^k_2 > inter_2$}{
\STATE 
$\ddot{M}_{i}^u \leftarrow \ddot{M}_{i}^u \cup \widetilde{M}_{i,:,k}^u$;}
\ENDIF
}
\ENDIF
\ENDFOR
\IF{$\ddot M_{i}^u == \emptyset$}
\STATE Replace the enhanced pseudo-label $\ddot{M}_{i}^u$ with the teacher's predicted pseudo-labels: $\ddot M_{i}^u \leftarrow \hat {M}_{i}^u$;
\ENDIF
\end{algorithmic}
\end{algorithm}
\subsubsection{Pixel-Wise Weighted Adjustment}
\label{PWLA}
Despite the effectiveness of our two strategies for refining pseudo-labels, there are cases where the scores do not exceed a certain threshold, indicating a mismatch between SAM-generated segments and the current pseudo-labels. In such situations, as inspired by previous work~\cite{yang2023semi}, we implement the Pixel-Wise Weighted Adjustment (PWA). PWA's core objective is to assign weights to pixels based on their confidence levels. High-confidence pixels, with scores near 0 or 1, indicate certainty in foreground or background prediction and are given higher weights. In contrast, pixels with scores around 0.5, often associated with noise or ambiguity, receive lower weights to reduce their influence on training. The mapping function $\Psi$ for translating pixel confidence into weights is defined as:
% \vspace{-0.8em}
\begin{equation}
\Psi(\Hat M_{i,j}^u)=\gamma-\frac{1}{\sqrt{2 \pi} \sigma} \exp \left(-\frac{(\hat M_{i,j}^u-\mu)^2}{2 \sigma^2}\right),
\label{eq:wbce}
% \vspace{-0.8em}
\end{equation}
\noindent where $\gamma,\sigma^2,\mu$ are hyperparameters, which are set to 1.3, 0.1, and 0.5 respectively. \par
Therefore, the loss for $i$-th unlabeled image $\mathcal{I}_i^u$ is defined as follows:
\begin{small}
\begin{equation}
\mathcal{L}_{unsup}=\frac{1}{H \times W}  \sum_{j=1}^{H \times W} \Psi(\Hat M_{i,j}^u) * L_{BCE}\left(M_{i,j}^u, \hat{M}_{i,j}^u\right).
\label{loss:pixel-bce}
\end{equation} 
\end{small}

\section{Experiment}
\begin{table*}[t]
% \small
\centering
\setlength{\tabcolsep}{5pt}
% \vspace{-0.5em
\caption{\textbf{Comparison of supervised, baseline and our proposed SemiRES on RefCOCO, RefCOCO+ and G-Ref.} For all approaches, we use LAVT~\cite{yang2022lavt} as the RES model.  ``Supervised'' denotes the fully-supervised training with only labeled data. ``Baseline'' denotes the plain semi-supervised training with data augmentation using both labeled and unlabeled data.}
\begin{tabular}{lccccccccccccccc}
\toprule[1.2pt]
\multirow{2}{*}{Methods} & \multicolumn{15}{c}{\textit{\textbf{RefCOCO}}}          \\ \cline{2-16} 
         & \multicolumn{3}{c}{0.5\%}          &      & \multicolumn{3}{c}{1\%}            &      & \multicolumn{3}{c}{2\%}            &      & \multicolumn{3}{c}{5\%}           \\ \cline{1-4} \cline{6-8} \cline{10-12} \cline{14-16} 
         & val  & testA& testB&      & val  & testA& testB&      & val  & testA& testB&      & val  & testA& testB\\ \cline{2-4} \cline{6-8} \cline{10-12} \cline{14-16} 
Supervised                  
    & 22.37
    & 25.35 
    & 19.28
    &
    & 32.26 
    & 35.71  
    & 28.02
    &
    & 39.46
    & 42.50
    & 35.26
    &
    & 49.40 
    & 53.72 
    & 44.87\\
Baseline        
    & 30.33
    & 35.18
    & 25.74
    &   
    & 42.62 
    & 48.86 
    & 37.43 
    &  
    & 51.05
    & 54.75
    & 46.34
    &  
    & 60.58  
    & 64.98  
    & 54.85 \\
SemiRES
    &  40.31
    &  46.48
    &  34.88
    &   
    & 50.90 
    & 57.54
    & 44.48 %\scriptsize{\textcolor{blue}{(+7.20)}} 
    &  
    &  54.85
    & 60.39
    & 48.52
    & 
    & 61.31
    & 66.64
    & 55.94 \\
    \midrule[0.8pt]
 \midrule[0.8pt]
 
\multirow{2}{*}{Methods} & \multicolumn{15}{c}{\textit{\textbf{RefCOCO+}}}          \\ \cline{2-16} 
         & \multicolumn{3}{c}{0.5\%}          &      & \multicolumn{3}{c}{1\%}            &      & \multicolumn{3}{c}{2\%}            &      & \multicolumn{3}{c}{5\%}           \\ \cline{1-4} \cline{6-8} \cline{10-12} \cline{14-16} 
         & val  & testA& testB&      & val  & testA& testB&      & val  & testA& testB&      & val  & testA& testB\\ \cline{2-4} \cline{6-8} \cline{10-12} \cline{14-16} 
Supervised                  
    & 20.83
    & 24.53 
    & 16.25
    &
    & 24.76 
    & 29.11  
    & 20.29
    &
    & 28.88
    & 32.41
    & 24.49
    &
    & 37.60
    & 42.32 
    & 32.39\\
Baseline        
    &  25.89
    &  30.42
    &  20.23
    &   
    &  30.91
    &  35.83
    &  24.98  
    &
    &  36.98
    &  41.44
    &  30.63
    &  
    &  46.29
    &  52.46
    &  38.61\\
SemiRES
    &  31.99
    &  38.06
    &  25.92
    &   
    & 36.49
    & 42.86
    & 28.58
    &  
    & 40.41
    & 46.84  
    & 33.30
    &
    &  47.00
    & 54.42
    & 38.74
    \\ \midrule[0.8pt]
 \midrule[0.8pt]
 
\multirow{2}{*}{Methods} & \multicolumn{15}{c}{\textit{\textbf{G-Ref}}}          \\ \cline{2-16} 
         & \multicolumn{3}{c}{0.5\%}          &      & \multicolumn{3}{c}{1\%}            &      & \multicolumn{3}{c}{2\%}            &      & \multicolumn{3}{c}{5\%}           \\ \cline{1-4} \cline{6-8} \cline{10-12} \cline{14-16} 
         & val(U)& \multicolumn{2}{c}{test(U)}  &      & val(U)& \multicolumn{2}{c}{test(U)}  &      & val(U)& \multicolumn{2}{c}{test(U)}  &      & val(U)& \multicolumn{2}{c}{test(U)}  \\ \cline{2-4} \cline{6-8} \cline{10-12} \cline{14-16} 
Supervised                  
    &  18.33
    & \multicolumn{2}{c}{18.69 }
    &  
    &  24.31
    & \multicolumn{2}{c}{24.72}
    &  
    & 28.23
    & \multicolumn{2}{c}{29.86}
    &   
    & 37.25
    & \multicolumn{2}{c}{38.62}
    \\
Baseline        
    &  26.02
    &  \multicolumn{2}{c}{27.62}
    &   
    &   30.91
    &  \multicolumn{2}{c}{31.51}  
    &  
    &  37.07
    &  \multicolumn{2}{c}{38.55}
    &    
    &  46.67
    & \multicolumn{2}{c}{48.39}
    \\
SemiRES
    &  31.81
    &  \multicolumn{2}{c}{33.40}
    &   
    &  34.76
    &  \multicolumn{2}{c}{36.18}  
    &  
    &  42.15
    &  \multicolumn{2}{c}{43.49}
    &  
    &  47.61
    & \multicolumn{2}{c}{50.11}
    \\
    \bottomrule[1.2pt]
\end{tabular}
% \vspace{-1em}
\label{tab:total_new}
\end{table*}

\subsection{Datasets}
We verify the effectiveness of our proposed method on three standard RES benchmark datasets, RefCOCO~\cite{yu2016modeling}, RefCOCO+~\cite{yu2016modeling}, and G-Ref~\cite{mao2016generation,nagaraja2016modeling}.
Images in these datasets are collected from the MS-COCO dataset~\cite{lin2014microsoft} and are attached with one or more short captions.\\
% These datasets comprise images collected from the MS COCO dataset~\cite{lin2014microsoft}, each annotated with natural language expressions.
% 2023 11 2 23:35
\textbf{RefCOCO \& RefCOCO+} contains 19,994, 19,992 images, with
50,000, 49,856 annotated objects and 142,209, 141,564 annotated expressions, respectively. RefCOCO and RefCOCO+ are split into four parts, \emph{i.e.}, train, val, testA and testB. The expressions of RefCOCO are mainly about absolute position, while the ones of RefCOCO+ includes more information related to attributes.\par
\noindent\textbf{G-Ref} contains 26,711 images, with 54,822 annotated objects and 104,560 annotated expressions. In contrast, G-Ref contains more intricate expressions, with an average length of 8.4 words, making the dataset more challenging. Moreover, the G-Ref dataset is split into two distinct partitions, one maintained by UMD and the other by Google, and we present results for UMD split.
\subsection{Implementation Details}
Our experimental setup uses LAVT~\cite{yang2022lavt} as the baseline for the RES network, employing the same Swin Transformer~\cite{liu2021swin} and BERT~\cite{devlin2018bert} backbones for visual and linguistic modalities, respectively. We implement our SemiRES model in PyTorch~\cite{paszke2019pytorch}, training it on 4 RTX3090 GPUs with 3 labeled and 3 unlabeled samples per GPU. Optimization is done using the AdamW optimizer, with an initial learning rate of $5\times 10^{-5}$ and weight decay of $10^{-2}$. Data augmentation includes RandomColorJitter and RandomGaussianBlur. We set the EMA rate at 0.996 and use pre-trained weights of the ViT-Huge version for SAM in generating multi-scale masks.\par
We use the overall Intersection-over-Union (oIoU) metric~\cite{ding2021vision,yang2022lavt,liu2023polyformer}, a standard in RES, to measure the overlap ratio between predicted masks and ground truth. % by calculating the intersection area divided by the union area.

\subsection{Experimental Results}
\subsubsection{Comparison with supervised model and baseline}
In Tab.~\ref{tab:total_new}, we conduct experiments on RefCOCO, RefCOCO+ and G-Ref under the setting of 0.5\%, 1\%, 2\% and 5\% labeled data. From the results, it can be observed that the performance of the supervised model dramatically drops when lacking sufficient labeled data. For instance, with 0.5\% labeled data, the overall IoU on RefCOCO val set is only 22.37\%. We also compare the plain semi-supervised baseline, as mentioned in Sec.~\ref{baseline}, which surpasses the supervised method in all settings, \emph{i.e.}, exhibiting a +7.96\% improvement on RefCOCO val set, with 0.5\% labeled data. Most importantly, our proposed SemiRES achieves state-of-the-art performance compared to baseline. In comparison to the supervised model, SemiRES gains +17.94\%, +18.64\%, +15.39\%, and +11.91\% on RefCOCO val set under the setting of 0.5\%, 1\%, 2\%, and 5\% labeled data, respectively.
\subsubsection{Ablation Study}
To validate the effectiveness of components in SemiRES, we conduct the ablation study on the 1\% labeled data and the remaining 99\% unlabeled data on RefCOCO.\par
\noindent \textbf{The comparison of different matching strategies.}
In Tab.~\ref{tab:ablation_matching}, we evaluate our two proposed matching strategies: IOM and CPI. Both strategies significantly surpass the baseline, demonstrating their effectiveness. IOM, which matches the top-1 IoU mask from SAM, attains 49.66\% oIoU on the RefCOCO val set. This result not only indicates the simplicity and efficacy of IOM but also corroborates SAM's exceptional segmentation ability, capable of producing ideal masks in most instances. However, IOM is slightly less effective compared to CPI, particularly when pseudo-labels segment only a small part of the target.\par
CPI-O, tailored for over-segmentation scenarios, effectively eliminates noise by filtering out excessively segmented small regions. Nonetheless, its performance increment is less pronounced (45.23\% vs. 42.62\%) when the noisy region enlarges and starts matching with new noisy areas. In contrast, CPI-U, designed to tackle under-segmentation, emerges as the most performant strategy (50.90\% vs. 42.62\%). This superior performance of CPI-U can be attributed to its efficient resolution of the common under-segmentation problem in pseudo-labels. When CPI-O and CPI-U are combined into a CPI strategy, there is a minor decrease in performance, likely due to CPI-U inadvertently introducing noise in the pseudo-label refinement process. \par
\begin{table}[t]
% \small
\centering
\vspace{-5pt}
\caption{Ablation study of SAM Matching Strategy.}
\begin{tabular}{@{}c|ccc@{}}
\toprule
%  Random Gaussian filter(GF)
% Random horizontal flipping(HF)
% crop(C)\\
% Color jittering(CJ)
SAM Matching Strategy & val & testA & testB  \\ \midrule
baseline&42.62&48.86&37.43\\ \hline
% \rowcolor{gray!18} % 设置颜色为浅灰色
% IOM&50.70&57.27&	44.20\\
% \rowcolor{gray!18} % 设置颜色为浅灰色
IOM&49.66&55.26&44.09	\\ \hline
% DAM w/o rle&47.67 &53.27 &43.84\\ \hline
% DAM*&47.67 &53.27 &43.84\\ \hline
% DAM-left&44.60&50.81&	38.09\\
CPI-O&45.23 &52.53&38.30	\\
\rowcolor{gray!18} % 设置颜色为浅灰色
% DAM-right&50.76 &57.34 &44.87	\\
% \rowcolor{gray!18} % 设置颜色为浅灰色
CPI-U&\textbf{50.90} &\textbf{57.54} &44.48	\\
%DAM-right& & &	\\
CPI &50.37 &56.88 &\textbf{44.52}	\\ 
\bottomrule
\end{tabular}
% \vspace{-1em}
\label{tab:ablation_matching}
\end{table}

% \vspace{-0.5em}
\begin{figure*}[h]
  \centering
\includegraphics[width=0.95\linewidth]{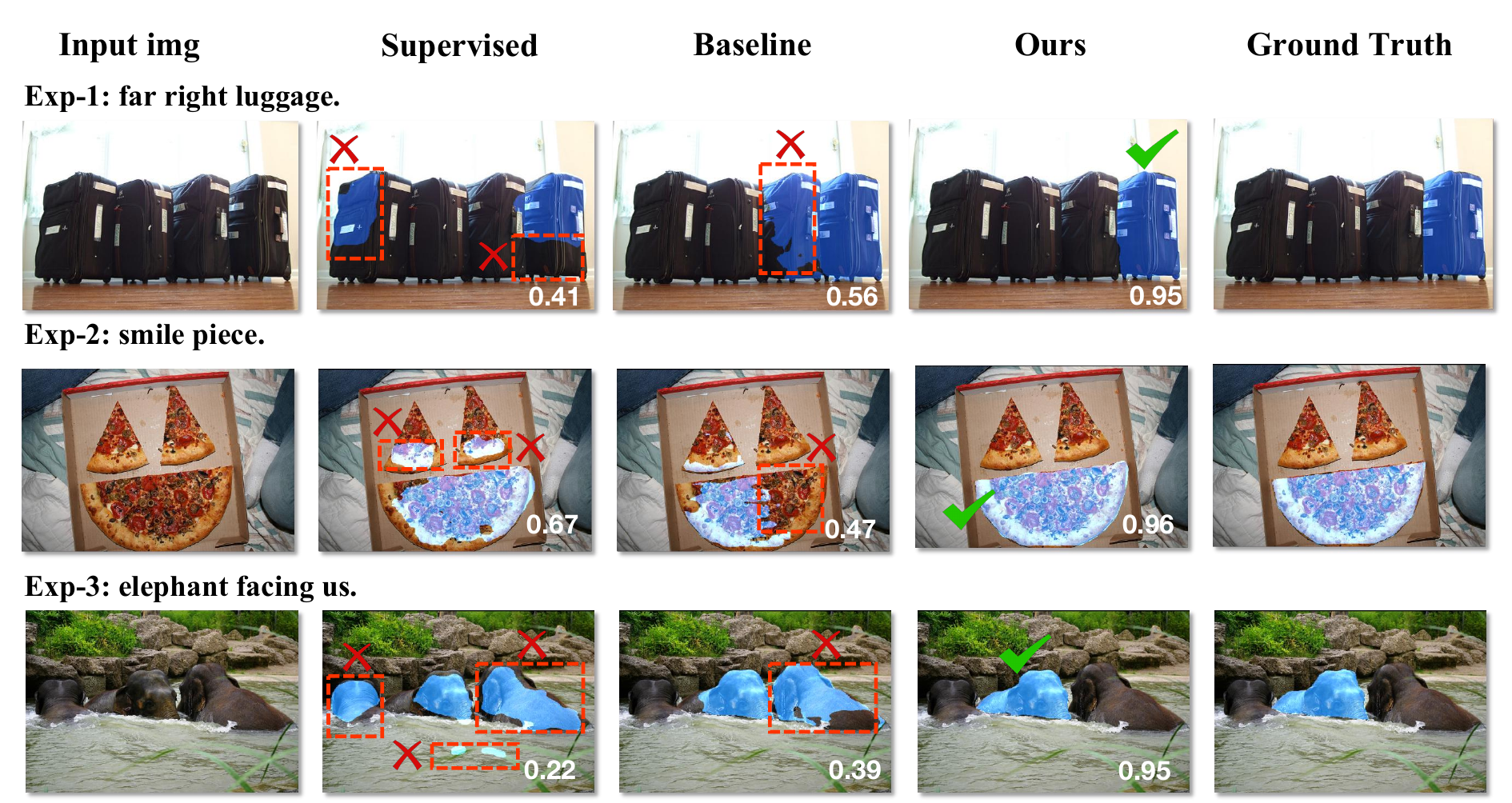}
  %\vspace{-10pt}
  \caption{\textbf{Qualitative analysis for SemiRES, supervised model, semi-supervised baseline and ground truth.} The white number in the bottom right corner represents the IoU value between predicted image and ground truth. The object enclosed by the red dashed box represents incorrect segmentation. Here we use supervised and semi-supervised model trained on 1\% labeled data for visualization.}
  \label{fig:visualization}
  \vspace{-1em}
\end{figure*}
\begin{table}[t]
% \small
% \vspace{-5pt}
\centering
\caption{Ablation study of different thresholds for matching.}
\begin{tabular}{@{}ccc|ccc@{}}
\toprule
%  Random Gaussian filter(GF)
% Random horizontal flipping(HF)
% crop(C)\\
% Color jittering(CJ)
$inter_1$ & $inter_2$&$IoU_{rate}$ & val & testA & testB  \\ \midrule
-&-&0.7& 48.24 & 55.34 &41.83 \\
-&-&0.6& 49.36 & 56.35 &42.94\\
-&-&0.5& 49.66 & 55.26 &44.09\\  \hline
0.7&0.7&-& 50.37 &56.88 &\textbf{44.52}\\
0.6&0.6&-& 50.18 & 56.97 & 44.51\\
0.5&0.5&-& 49.38 & 56.40 &43.07 \\ \hline
\rowcolor{gray!20} % 设置背景颜色为浅灰色
0.7&-&-&\textbf{50.90}& 57.54& 44.48\\
0.6&-&-&50.13 &\textbf{57.66}  &  43.71\\
0.5&-&-& 50.21 & 57.03 &44.17  \\
\bottomrule
\end{tabular}
% \vspace{-1em}
\label{tab:ablation_threshold}
% \vspace{-1em}
\end{table}
\begin{table}[t]
\centering
% \small
% \vspace{-5pt}
\caption{Comparison of SemiRES and filtering-based method.}
\begin{tabular}{@{}c|ccc@{}}
\toprule
Semi-Supervised Settings &val & testA & testB  \\ \midrule
Supervised&32.26&35.71&28.02\\\hline 
baseline& 42.62 & 48.86  & 37.43\\ \hline 
+confidence filtering&42.99&48.78&35.96\\
\cellcolor{gray!18}{+SemiRES}  &\cellcolor{gray!18}\textbf{50.90} &\cellcolor{gray!18}\textbf{57.54}&\cellcolor{gray!18}\textbf{44.48 }\\
\bottomrule
\end{tabular}
% \vspace{-1em}
\label{tab:filter}
\end{table}
\begin{table}[t]
\centering
% \vspace{-5pt}
% \small
\caption{Ablation study on various components: MLT (mutual learning training), DA (data augmentation), PWA (pixel-wise adjustment), and Refine (SAM matching refinement).}
\begin{tabular}{@{}cccc|ccccc@{}}
\toprule
 MLT & DA& PWA &Refine& val & testA & testB  \\ \midrule
\ding{55}&\ding{55}&\ding{55}&\ding{55} & 32.26
 &35.71 & 28.02\\
\ding{51}&\ding{55}&\ding{55}&\ding{55} & 40.84
 & 46.46 & 35.22 \\
\ding{51}&\ding{51}&\ding{55}&\ding{55} & 42.62 & 48.86& 37.43  \\
\ding{51}&\ding{51}&\ding{51}&\ding{55} & 43.09 &49.72 & 37.10 \\
\ding{51}&\ding{51}&\ding{55}&\ding{51} & 49.96 & 57.12&43.86  \\
\rowcolor{gray!20} % 设置背景颜色为浅灰色
\ding{51}&\ding{51}&\ding{51}&\ding{51} & \textbf{50.90} & \textbf{57.54} & \textbf{44.48} 
\\ \bottomrule
\end{tabular}
\vspace{-0.5em}
\label{tab:ablation_component}
\end{table}
\noindent \textbf{The impact of the matching threshold.}
In the proposed SemiRES, we use specific thresholds to optimize the matching rates for the IOM and CPI strategies. 
Our ablation study, shown in Tab.~\ref{tab:ablation_threshold}, reveals that IOM achieves its best performance with an $IoU_{rate}$ of 0.5. %This indicates that setting a high threshold for IOM may make it struggle to match the appropriate segment produced by SAM under such conditions. 
For CPI, a higher performance is observed when $inter_1$ and $inter_2$ are both set to 0.7. This setting is crucial as lower thresholds in CPI could include noise, especially in the CPI-U variant, where $inter_1=0.7$ yields the best results. By default, unless specified otherwise, $IoU_{rate}$, $inter_1$, and $inter_2$ are set to 0.5, 0.7, and 0.7, respectively.\par
\noindent \textbf{Comparison with filtering-based method.}
In our analysis, as presented in Tab.~\ref{tab:filter}, we compare SemiRES with a confidence filtering method that eliminates the lowest 5\% of pseudo-labels based on confidence scores. SemiRES demonstrates superior performance, achieving 50.90\% versus 42.99\% on the 1\% RefCOCO validation set. This result suggests that the filtering-based method is overly rigid, leading to suboptimal use of pseudo-labels.\par
\noindent \textbf{Effectiveness of different Components.}
We present our experimental analysis in Tab.~\ref{tab:ablation_component}, systematically ablated to evaluate each component of SemiRES. The first row uses only 1\% labeled data for supervised training. The second row adds the remaining 99\% as unlabeled data for basic semi-supervised learning. The third and fourth rows include data augmentation and the PWA module, demonstrating their effectiveness with incremental improvements (40.84\% vs. 42.62\% vs. 43.09\%). The fifth row shows a significant gain from using SAM for pseudo-label refinement, with a 7.34\% increase (49.96\% vs. 42.62\%). Finally, the sixth row indicates that PWA continues to improve performance even when SAM candidates do not match the pseudo-labels, highlighting the synergistic effect of these modules.\par
% \noindent\textbf{The Impact on Different RES Frameworks.}
% To assess the generalizability of SemiRES, we integrate it with the region-based GRES baseline ReLA~\cite{liu2023gres}. As Table~\ref{tab:res_baseline} shows, SemiRES significantly enhances performance compared to the ``Supervised'' approach (54.26\% vs. 30.95\%), demonstrating its robust generalization capabilities across different RES frameworks.
\subsection{Qualitative Analysis}
We showcase qualitative results in Fig.~\ref{fig:visualization}, comparing SemiRES with a supervised model, a semi-supervised baseline, and ground truth. Impressively, SemiRES corrects errors from both the supervised model and the semi-supervised baseline. For instance, in the first example, while the supervised and baseline models fail to interpret ``far right'' correctly, leading to inaccurate identification of the luggage, SemiRES precisely localizes the target. In the second example, SemiRES effectively understands ``smile'' and accurately segments the correct pizza. In a more complex third scenario with several elephants, SemiRES successfully identifies the elephant facing towards us, demonstrating its advanced understanding.

\section{Conclusion}
In this work, we present a novel semi-supervised framework, namely SemiRES, to address the challenge of costly annotations in RES. SemiRES incorporates two innovative matching strategies that leverage the robust segmentation capabilities of SAM to refine the quality of pseudo-labels. In situations where SAM is unable to rectify the pseudo-labels, we employ the Pixel-Wise Adjustment (PWA) strategy, which utilizes the original pseudo-labels for efficient training directly. Our extensive experiments demonstrate that SemiRES achieves competitive results on three RES benchmark datasets, underscoring its viability and effectiveness for real-world applications.
\section*{Impact Statement}
This paper presents work whose goal is to advance the field of Machine Learning. There are many potential societal consequences of our work, none which we feel must be specifically highlighted here.

\section*{Acknowledgements}
This work was supported by National Key R\&D Program of China (No.2023YFB4502804), the National Science Fund for Distinguished Young Scholars (No.62025603), the National Natural Science Foundation of China (No. U21B2037, No. U22B2051, No. 62072389), the National Natural Science Fund for Young Scholars of China (No. 62302411), China Postdoctoral Science Foundation (No. 2023M732948), the Natural Science Foundation of Fujian Province of China (No.2021J01002,  No.2022J06001), and partially sponsored by CCF-NetEase ThunderFire Innovation Research Funding (NO. CCF-Netease 202301).

\nocite{langley00}

\bibliography{example_paper}
\bibliographystyle{icml2024}

%%%%%%%%%%%%%%%%%%%%%%%%%%%%%%%%%%%%%%%%%%%%%%%%%%%%%%%%%%%%%%%%%%%%%%%%%%%%%%%
%%%%%%%%%%%%%%%%%%%%%%%%%%%%%%%%%%%%%%%%%%%%%%%%%%%%%%%%%%%%%%%%%%%%%%%%%%%%%%%
% APPENDIX
%%%%%%%%%%%%%%%%%%%%%%%%%%%%%%%%%%%%%%%%%%%%%%%%%%%%%%%%%%%%%%%%%%%%%%%%%%%%%%%
%%%%%%%%%%%%%%%%%%%%%%%%%%%%%%%%%%%%%%%%%%%%%%%%%%%%%%%%%%%%%%%%%%%%%%%%%%%%%%%
\newpage
\appendix
% \onecolumn

%%%%%%%%%%%%%%%%%%%%%%%%%%%%%%%%%%%%%%%%%%%%%%%%%%%%%%%%%%%%%%%%%%%%%%%%%%%%%%%
%%%%%%%%%%%%%%%%%%%%%%%%%%%%%%%%%%%%%%%%%%%%%%%%%%%%%%%%%%%%%%%%%%%%%%%%%%%%%%%
\section{The Labeling Cost at Different Ratios }
In semi-supervised learning, the more labeled data there is, the higher the cost of annotation. The three RES benchmark datasets RefCOCO, RefCOCO+, G-Ref contain 50000, 49856 and 54822 annotated objects, respectively. Following the label budget calculation in \cite{kim2023devil}
, manually labeling a mask for one instance takes approximately 79.1 seconds. Therefore, it can be observed that annotating masks for the entire dataset is a very time-consuming task. Based on this, we calculate the required annotation time for labeling 0.5\%, 1\%, 2\%, 5\%, 10\%, 20\%, and 50\% of the data on the RefCOCO as follows:
\begin{itemize}
\item 0.5\%:  50000  × 0.005 × 79.1 / 60 / 60 / 24 = 0.2 day
\item 1\%: 50000  × 0.01 × 79.1 / 60 / 60 / 24 = 0.5 day
\item 2\%: 50000  × 0.02 × 79.1 / 60 / 60 / 24 = 1.0 day
\item 5\%: 50000  × 0.05 × 79.1 / 60 / 60 / 24 = 2.3 day
\item 10\%: 50000  × 0.1 × 79.1 / 60 / 60 / 24 = 4.6 day
\item 20\%: 50000  × 0.2 × 79.1 / 60 / 60 / 24 = 9.2 day
\item 50\%: 50000  × 0.5 × 79.1 / 60 / 60 / 24 = 22.9 day
\item Fully-supervised: 50000 × 79.1 / 60 / 60 / 24 = 45.8 day
\end{itemize}

\section{The Impact of the Proportion of Labeled Data}
We further conduct the experiment using our SemiRES framework with a larger proportion of labeled data under the settings of 10\%, 20\%, 30\%, 40\%, and 50\% of labeled data, as shown in Tab.~\ref{tab:fraction}. We observe that when the labeled data approaches 30\%, the performance of our method nearly matches that of fully supervised models. This validates that our proposed SemiRES maintains great segmentation performance with a significant reduction in annotation costs.

\section{The Impact on Different RES Frameworks}
To assess the generalizability of SemiRES, we integrate it with the region-based GRES baseline ReLA~\cite{liu2023gres}. As Tab.~\ref{tab:res_baseline} shows, SemiRES significantly enhances performance compared to the ``Supervised'' approach (43.57\% vs. 31.54\%), demonstrating its robust generalization capabilities across different RES frameworks.

\begin{table}[t]
\centering
\setlength{\tabcolsep}{5pt}
\caption{The impact of the proportion of labeled data.}
\resizebox{0.48\textwidth}{!}{
\begin{tabular}{@{}cc|ccc@{}}
\toprule
Setting & Labeled Fraction &val & testA & testB  \\ \midrule
\multirow{5}{*}{SemiRES}&10\% & 63.60 & 68.43  & 60.20\\ 
&20\% & 65.97 & 69.15  & 62.08\\ 
&30\% & 68.54 & 70.62  & 63.90\\  
&40\% & 68.76 & 71.69  & 65.23\\ 
&50\% & 69.45 & 72.94  & 65.68 \\ \hline 
Fully-Supervised&-&72.73& 75.82& 68.79\\
\bottomrule
\end{tabular}}
\vspace{-1em}
\label{tab:fraction}
\end{table}
\begin{table}[t]
\small
\centering
\caption{Results of different state-of-the-art (SOTA) RES approaches equipped with the SemiRES strategy.}
\resizebox{0.48\textwidth}{!}{
\begin{tabular}{@{}cc|ccc@{}}
\toprule
RES Baseline & Settings &val & testA & testB  \\ \midrule
\multirow{2}{*}{LAVT~\cite{yang2022lavt}}&Supervised&32.26&35.71&28.02\\
&Baseline&42.62&48.86&37.43\\
&\cellcolor{gray!18}{SemiRES}  &\cellcolor{gray!18}\textbf{50.90} &\cellcolor{gray!18}\textbf{57.54}&\cellcolor{gray!18}\textbf{44.48 }\\ \hline 
\multirow{2}{*}{GRES~\cite{liu2023gres}}&Supervised&31.54&35.93&25.37\\
&Baseline&38.65&43.53&34.63\\
&\cellcolor{gray!18}{SemiRES}  
&\cellcolor{gray!18}\textbf{43.57}
&\cellcolor{gray!18}\textbf{49.05}&\cellcolor{gray!18}\textbf{39.05} \\
\bottomrule
\end{tabular}}
% \vspace{-1em}
\label{tab:res_baseline}
\end{table}

\begin{figure*}[] 
\centering 
\includegraphics[width=1\textwidth]{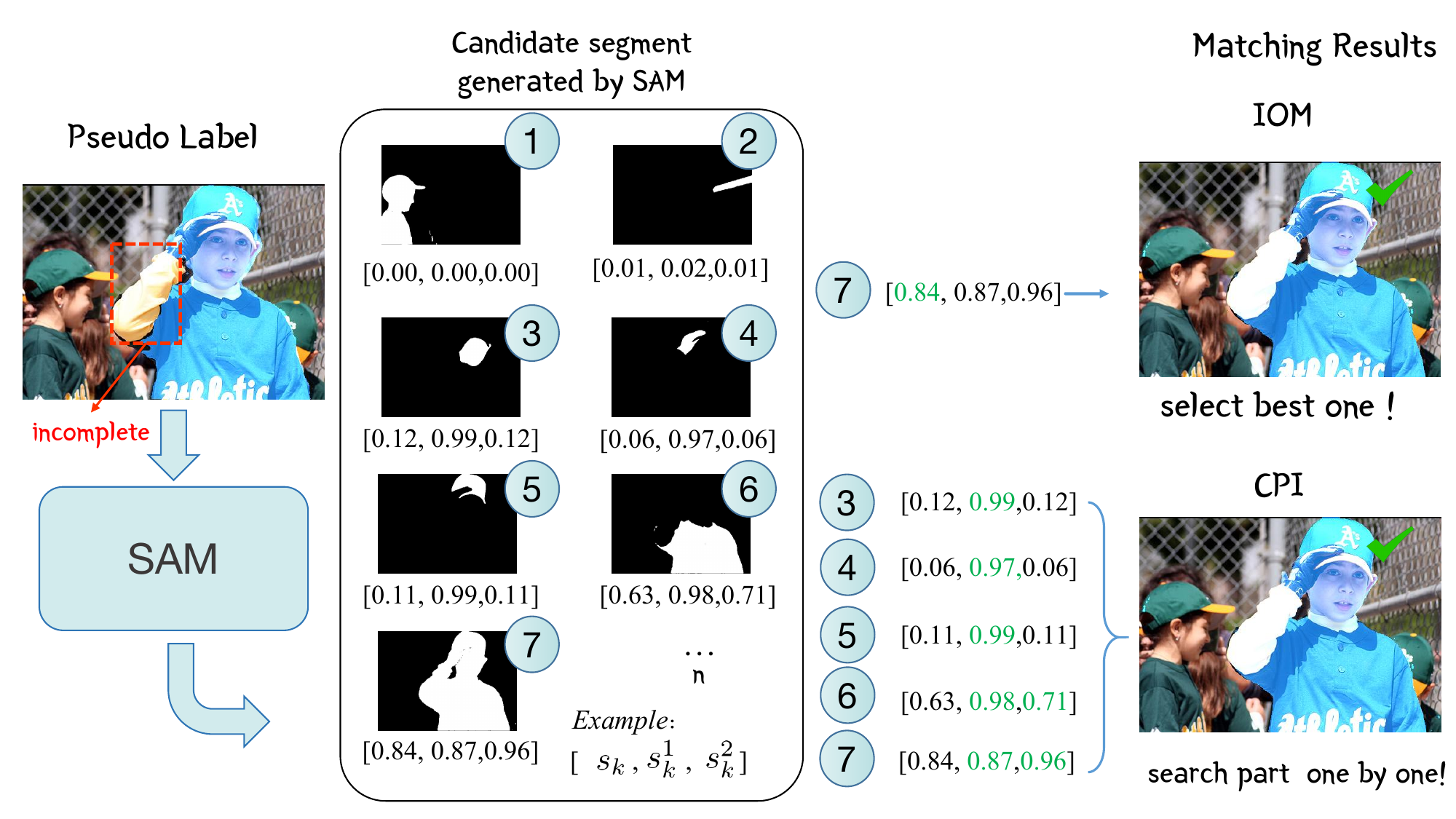} 
\caption{Demonstration of how IOM and CPI operate in matching pseudo-labels. The caption for this image is ``kid looking at you''.} 
\label{principal}
\end{figure*}
 
\begin{figure*}[t] 
\centering 
\includegraphics[width=1\textwidth]{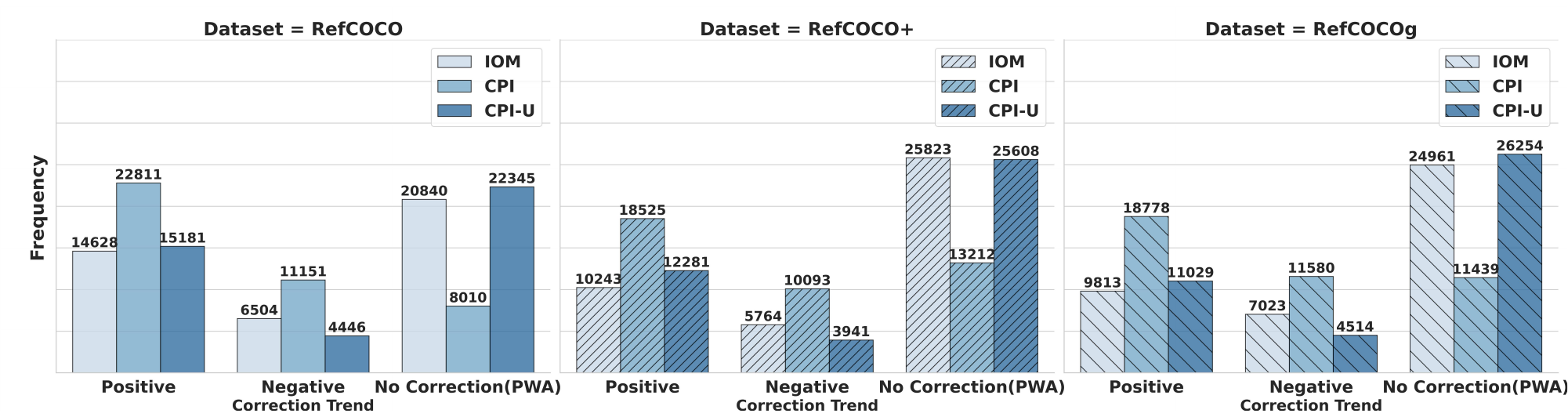} 
\caption{Quantitative statistics of positive, negative corrections and no corrections(PWA) for IOM, CPI and variants. }
\label{statistics}
\end{figure*}
\section{The Working Principles of IOM and CPI}
To provide a clearer understanding of the working principles underlying the two SAM-based Pseudo-Label Refinement modules we have introduced, namely the IOM and CPI method, as depicted in Fig.~\ref{principal}. Specifically, we select a sample to elucidate how IOM and CPI operate in matching pseudo-labels. Specifically, IOM selects the best-matching segment proposal that meets the criteria from the candidate pool within SAM and uses it as the refined pseudo-label. IOM algorithm selects the segment with index 7 as the optimal match to refine the pseudo-label. On the other hand, CPI leverages the multi-scale characteristics obtained from SAM segmentation, systematically selecting qualifying partial segments, and eventually merging them together to form the final result. For CPI, it selects all segments that meet the conditions, such as indices 3, 4, 5, 6, and 7. In the end, it takes the union of these segments as the final refined pseudo-label.
Observably, IOM and CPI, while both aimed at refining pseudo-labels within the SAM differ significantly in their operational approach and specialization. IOM is tailored for simpler, less noisy label refinement tasks, while CPI is designed to address more complex segmentation errors, together providing a comprehensive solution for enhancing the accuracy of pseudo-labels in segmentation tasks.
\begin{figure}[] 
\centering 
\includegraphics[width=0.9\columnwidth]{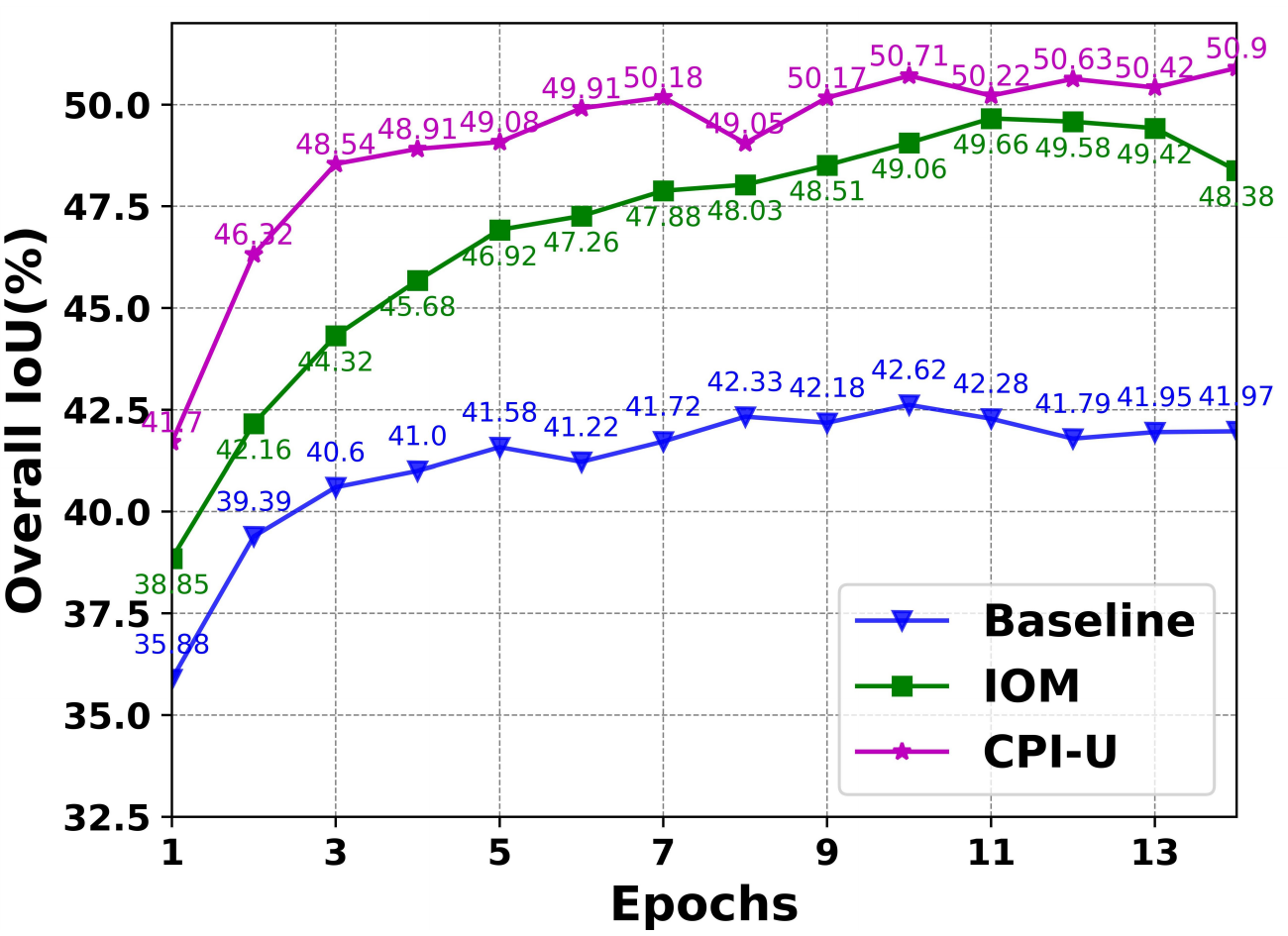}  
\vspace{-1em}
\caption{Training curves of two matching strategies of SemiRES and semi-supervised baselines.} 
\label{fig:overall_epoch}
\end{figure}
\section{Comparasion with other class-agnostic proposal network}
\begin{table}[t]
\centering
% \small
\vspace{-5pt}
\caption{Comparison with other class-agnostic
proposal network.}
\begin{tabular}{@{}c|ccc@{}}
\toprule
Semi-Supervised Settings &val & testA & testB  \\ \midrule
Supervised&32.26&35.71&28.02\\\hline 
SemiRES(+SEEM)&47.61&56.28&43.62\\
\cellcolor{gray!18}{SemiRES(+SAM)}  &\cellcolor{gray!18}\textbf{50.90} &\cellcolor{gray!18}\textbf{57.54}&\cellcolor{gray!18}\textbf{44.48 }\\
\bottomrule
\end{tabular}
\vspace{-1em}
\label{tab:seem}
\end{table}
To validate the tight combination between our SemiRES and SAM, we replace SAM with SEEM~\cite{zou2024segment} for generating class-agnostic mask proposals and applied our matching strategy to refine pseudo-labels. As shown in Tab.~\ref{tab:seem}, the results maintained under the same configurations as detailed in our paper, indicated that the performance did not match that achieved using SAM for proposal extraction. Our analysis suggests that this discrepancy stems from the designed CPI algorithm capitalizing on SAM's strong segmentation capability and its ability to extract multi-scale proposal masks, which facilitated effective pseudo-label optimization, a feat not typically achievable with general class-agnostic proposal networks.

\section{The potential of SemiRES for detecting small objects}
To explore the potential of our proposed SemiRES method for detecting small objects, we curated samples from the RefCOCO training, validation, and test sets, organizing them by the size of the ground truth masks from smallest to largest, and specifically selected the top 5\% and 10\% of samples featuring small objects. As shown in Tab.~\ref{tab:smallobject}, experiments conducted on these subsets serve to validate the efficacy of our SemiRES approach in comparison to traditional supervised methods when focusing on small objects. Our findings reveal that our method significantly outperforms the supervised approach in this area. This improvement is attributed to the SAM's ability to utilize a multi-scale library of offline-generated masks, which includes candidates for small objects, allowing for the refinement of pseudo-labels. 
\begin{table}[t]
\centering
\setlength{\tabcolsep}{5pt}
\vspace{-5pt}
\caption{ The potential of SemiRES for detecting small objects.}
\begin{tabular}{@{}cc|ccc@{}}
\toprule
Setting & Method &val & testA & testB  \\ \midrule
\multirow{2}{*}{Top 5\% small objects}&Supervised & 11.37 & 13.84  & 11.75\\ 
&SemiRES & 20.45 & 27.21  & 17.04\\  \hline 
\multirow{2}{*}{Top 10\% small objects}&Supervised & 13.70 & 14.88 & 12.59\\  
&SemiRES & 24.98 & 29.76  & 18.05\\ 
% \multirow{2}{*}{Top 20\% small objects}&Supervised & 15.43 & 16.80 & 13.37 \\ 
% &SemiRES&26.68& 75.82& 68.79\\
\bottomrule
\end{tabular}
\vspace{-1em}
\label{tab:smallobject}
\end{table}

\section{Quantitative Statistics for Pseudo-label Refinement}
Due to the presence of noise and incompleteness in pseudo-labels, our proposed two SAM-based Pseudo-Label Refinement modules effectively enhance the quality of pseudo-labels, facilitating the mutual learning process between teacher and student models. To further analyze the roles of the designed IOM and CPI, along with their variants, we numerically compute their frequency for positive, negative corrections and no correction of pseudo-labels, as illustrated in Fig.~\ref{statistics}. Here we use the ground truth of all unlabeled data in three RES datasets to validate the correction performance of our methods. Positive correction indicates that, after the matching refinement, the corrected pseudo-label has a higher IoU with the ground truth than the original pseudo-label, thereby improving the quality of the pseudo-label. Negative correction, on the other hand, is the opposite; the quality of the updated pseudo-label is worse, exacerbating the misguidance in student learning. No correction indicates that the matching algorithms do not find a suitable correction for the pseudo label, thus the Pixel-Wise Adjustment (PWA) strategy mentioned in this paper is used for adjustment. The histogram results show that, among the three datasets for RES, the CPI method performs best in positive correction, while its variant CPI-U has the least occurrence of negative correction. And CPI-U exhibits faster convergence as shown in Fig.~\ref{fig:overall_epoch}.

% \section{Visualization of Pseudo Heatmaps}
% To better showcase our model's improvement in pseudo-label quality, we visualize the predicted pixel-level pseudo heatmaps during the stepwise convergence process, as shown in Fig.~\ref{fig:heatmap}. It is evident from the visuals that our proposed SemiRES progressively enhances the quality of pseudo-labels.

\section{Additional Visualizations}
We present more comparative visualizations of our proposed SemiRES with both supervised and baseline models, alongside the ground truth, as illustrated in Fig.~\ref{fig:vis_more3}.
Through these extensive examples, it becomes apparent that our proposed SemiRES excels in understanding semantic attributes such as absolute positional terms, spatial relations, and colors. Moreover, it demonstrates a strong capability in comprehending the semantics of complex sentences. Furthermore, it adeptly handles intricate details, showcasing robust proficiency in managing nuanced information. These visualizations underscore the effectiveness of our proposed SemiRES in correcting noisy and incomplete pseudo-labels, significantly enhancing performance, especially in scenarios with limited data.
\begin{figure*}[h]
   \centering
   \includegraphics[width = 0.85\textwidth]{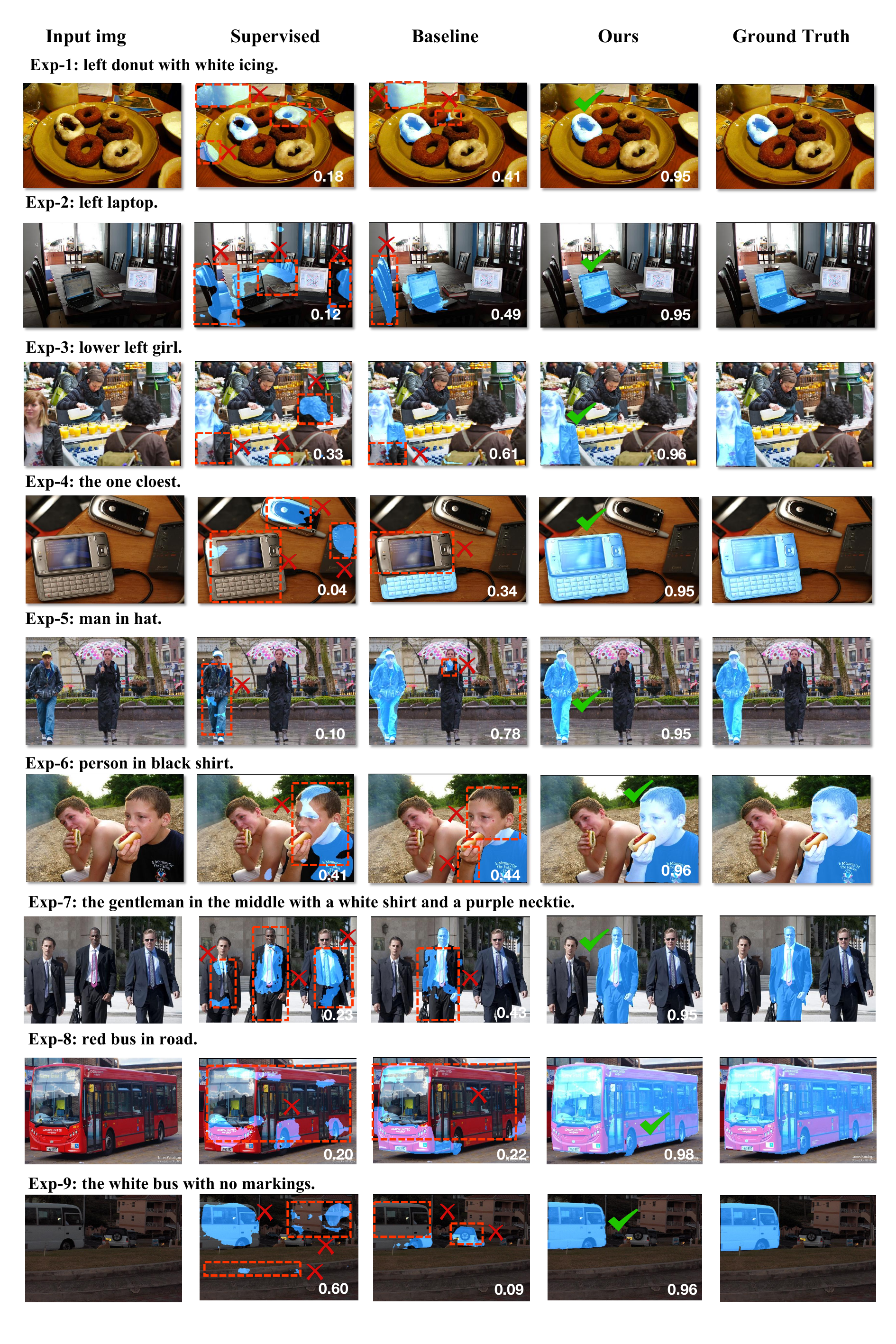}
  \label{fig:vis_more}
 \end{figure*}
 
\begin{figure*}[t]
   \centering
   \includegraphics[width = 0.85\textwidth]{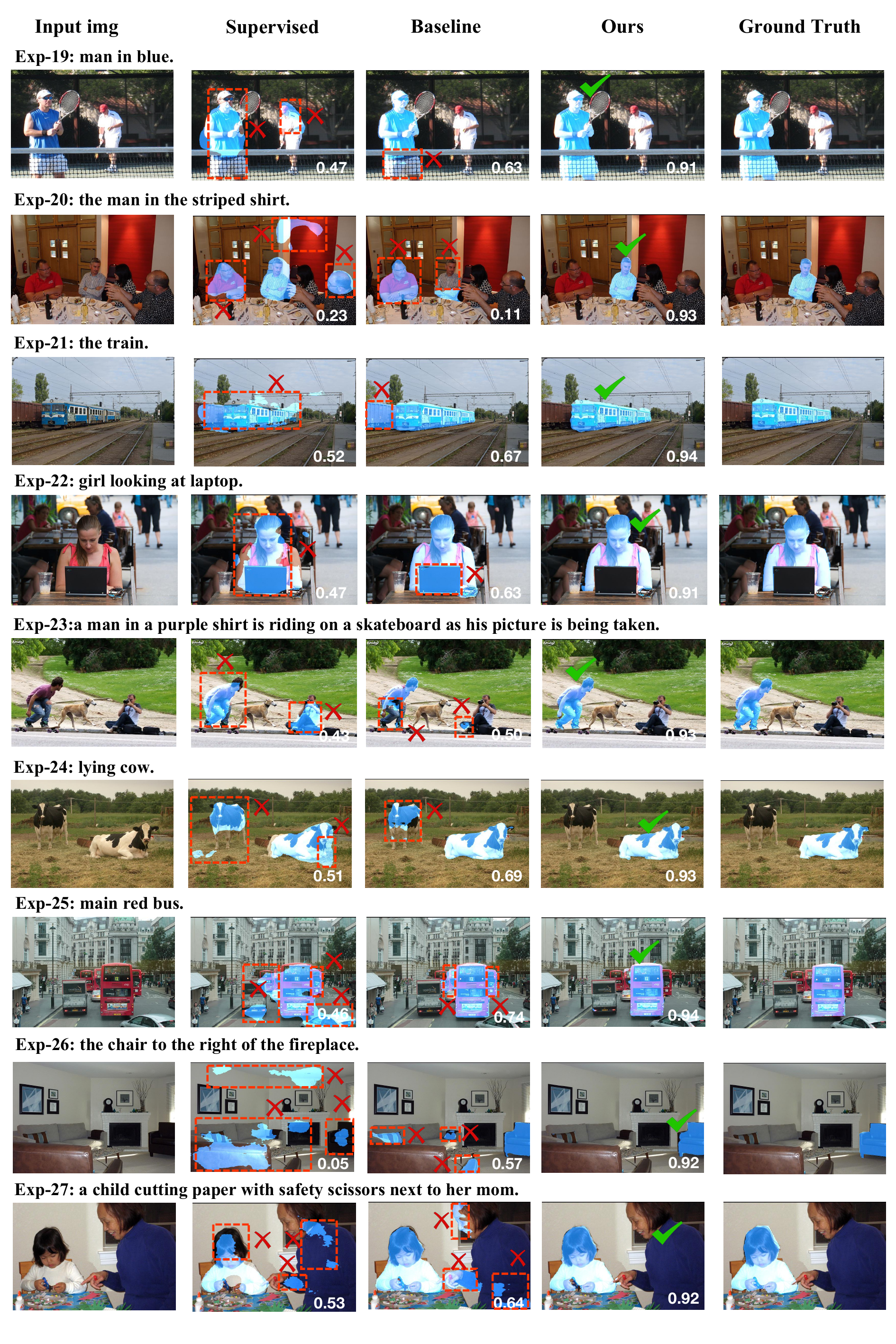}
   % \vspace{-1em}
   \caption{ More visualization results of our proposed SemiRES, compared with the supervised and baseline model. The red dashed bounding boxes denote regions where our model has made accurate predictions, while other models have made inaccurate predictions. }
  \label{fig:vis_more3}
 \end{figure*}

\end{document}